\documentclass{svproc}
\usepackage{url}
\usepackage{graphicx}
\usepackage{multicol}
\usepackage{footmisc}

\begin{document}
\mainmatter              
\title{Grey Level Texture Features for Segmentation of Chromogenic Dye RNAscope From Breast Cancer Tissue}
\titlerunning{Grey Level Texture Features for Segmentation of RNAscope}  
%
\author{Andrew Davidson*\inst{1} \and Arthur Morley-Bunker\inst{2} \and
George Wiggins\inst{2} \and Logan Walker\inst{2} \and Gavin Harris\inst{3} \and Ramakrishnan Mukundan\inst{1} \and
kConFab Investigators\inst{4,5}}
\authorrunning{Andrew Davidson et al.} 
%
\tocauthor{Andrew Davidson, Arthur Morley-Bunker, George Wiggin, Logan Walker, Gavin Harris, Ramakrishnan Mukundan, kConFab Investigators}
\institute{Department of Computer Science and Software Engineering, University of Canterbury, Christchurch, New Zealand\\
\email{andrew.davidson@pg.canterbury.ac.nz}
\and
Department of Pathology and Biomedical Science, University of Otago, Christchurch, New Zealand
\and
Canterbury Health Laboratories, Christchurch, New Zealand
\and
Sir Peter MacCallum Department of Oncology, The University of Melbourne, Melbourne, Victoria, Australia
\and
Peter MacCallum Cancer Center, Melbourne, Victoria, Australia
}

\maketitle              

\begin{abstract}
Chromogenic RNAscope dye and haematoxylin staining of cancer tissue facilitates diagnosis of the cancer type and subsequent treatment, and fits well into existing pathology workflows. However, manual quantification of the RNAscope transcripts (dots), which signify gene expression, is prohibitively time consuming. In addition, there is a lack of verified supporting methods for quantification and analysis. This paper investigates the usefulness of gray level texture features for automatically segmenting and classifying the positions of RNAscope transcripts from breast cancer tissue. Feature analysis showed that a small set of gray level features, including Gray Level Dependence Matrix and Neighbouring Gray Tone Difference Matrix features, were well suited for the task. The automated method performed similarly to expert annotators at identifying the positions of RNAscope transcripts, with an $F_1$-score of 0.571 compared to the expert inter-rater $F_1$-score of 0.596. These results demonstrate the potential of gray level texture features for automated quantification of RNAscope in the pathology workflow.
\end{abstract}
\section{Introduction}

In 2020, there were over 2.3 million new cases of breast cancer, and 685,000 deaths caused by breast cancer \cite{bc-increasing}. The effectiveness of breast cancer treatments vary based on some qualities of the cancer, therefore, it is crucial that these qualities are classified accurately and consistently. Over the last several years the number of cancer cases has increased, and the number of cases is projected to continue increasing \cite{bc-increasing}. Together with a shortage in the number of pathologists available \cite{pathologist_shortage}, these trends have led to an increased pathology workload. 

Within the last decade whole slide imaging has become common, which is the practice of using advanced digital scanners to produce high resolution images of patient tissue \cite{availability-tech,whole-slide-imaging}. There is an emerging opportunity to apply image processing methods to whole slide images, to produce consistent and exhaustive quantification of features relevant to breast cancer diagnosis. 

\subsection{RNAscope Staining}
RNAscope is an \textit{in situ} hybridization assay that is used to detect the presence of certain RNA sequences in tissue \cite{rnascope}. It allows the expression of specific gene sequences in a tissue sample to be visually quantified as stained dots (transcripts). The density of RNAscope transcripts is an indicator of the level of gene expression. This technique is applicable to the problem of breast cancer diagnosis because the ideal treatment differs based on some genetic characteristics of the tumour tissue, such as the level of expression of the Erb-B2 Receptor Tyrosine Kinase 2 (\textit{ERBB2}) gene, also referred to as \textit{HER2} status. 

To fit most easily into existing pathology workflows, a chromogenic RNAscope dye can be applied to tissue alongside haematoxylin, which is a stain that gives good definition to the nuclei. A stained sample can then be scanned on a slide in normal lab conditions to produce a single image. Fluorescent RNAscope dyes are also available and can be used to produce spectral image stacks (multiple images) that are much simpler to quantify RNAscope from. However, the fluorescent dyed samples deteriorate quickly, require a dark room to scan, and are more difficult to prepare. Therefore, this paper will solely investigate the segmentation of chromogenic RNAscope.

\subsection{RNAscope Segmentation}
\label{section:difficulties}
\begin{figure}[htbp]
    \centering
    \includegraphics[width=0.4\textwidth]{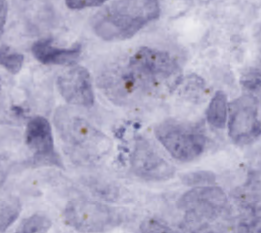}
    \caption{A small patch of haematoxylin and chromogenic RNAscope stained tissue, showing RNAscope transcripts with varying shape, size, and hue. Since the level of RNAscope staining is low, an accurate RNAscope segmentation is useful to determine the level of gene expression.}
    \label{fig:rnascope_single_var}
\end{figure}

The cases that are important to accurately segment are those with low or non-existent RNAscope staining. If there is heavy RNAscope staining, the staining is readily evident and gene expression can be deduced without having to count the individual transcripts. However, there are some challenges involved with designing a robust chromogenic RNAscope segmentation method. Since the RNAscope transcripts present at varying hue, stain intensity, and shape depending on stain preparation and tissue characteristics, segmentation is not straightforward; simple colour filtering is inadequate unless the preparation is consistently impeccable. This variability can be seen in figure \ref{fig:rnascope_single_var}. There is also a lack of available data with this stain type that also has annotations for the position of each RNAscope transcript. 

This paper details the development and testing of a gray level texture feature based RNAscope segmentation method, which operates on data from whole slide images. The images contain FFPE (formalin fixed paraffin embedded) breast cancer tissue that has been stained with haematoxylin and chromogenic RNAscope.

\section{Related Work}
\label{section:lit-review}
Gray level texture features are useful descriptors that are commonly used to extract information from medical images \cite{gray-level-used}. Gray level texture feature extraction techniques include the gray level co-occurrence matrix \cite{GLCM} (GLCM), which assesses the co-occurence of gray values with a given offset; the gray level run length matrix \cite{GLRLM} (GLRLM), which assesses runs of same valued consecutive pixels, and the gray level size zone matrix \cite{GLSZM} (GLSZM), which assesses connected zones of same valued pixels. Further methods include the gray level dependence matrix \cite{GLDM} (GLDM), which assesses the frequency of near, similar valued pixels at each gray level, and the neighbouring gray tone difference matrix \cite{NGTDM} (NGTDM), which assesses the variability of gray values from the average value of their nearby pixels.

A recent study \cite{MORLEYBUNKER2021151765} compared existing chromogenic RNAscope transcript counting methods on FFPE haematoxylin stained colorectal cancer tissues. Although it does not directly compare segmentation performance, this study indicates that currently available open source methods for RNAscope segmentation are slow due to requiring user input to function; by manually selecting the RNAscope positions or validating each RNAscope candidate. The only listed open-source method that does not require this level of user input (Trainable WEKA Segmentation) is not able to differentiate between single RNAscope transcripts and clusters of them, and is not considered fully automated due to the number of user steps required. Even the commercial methods considered (SpotStudio and Aperio RNA ISH) required some configuration to run, and did not perform better than the open source methods by most metrics assessed.

Although there are existing solutions, the methods published in academic literature focus on the clinical side of the problem and have not published accuracy metrics for their solutions. There are also few existing solutions that use haematoxylin and chromogenic RNAscope stained tissue, which is much more convenient to fit into the pathology workflow. Some commercial methods exist, but they do not have a verified accuracy, and the mechanisms by which they operate are not in the public domain. 

\section{Method}
\label{section:feature}
A texture feature-based method for RNAscope transcript segmentation was explored to assess the viability of this approach. It was implemented in Python, making use of the pyradiomics \cite{pyradiomics} library. 

\subsection{Data Acquisition}
Fourty whole slide images were obtained from the University of Otago, Christchurch. Each of these whole slide images contain a scan of a tissue microarray of haematoxylin and brown chromogenic RNAscope stained FFPE breast cancer tissues. The tissue was scanned at 40x magnification (0.25 microns per pixel).

\subsection{Expert Annotation of RNAscope Transcripts}
\label{section:expert_annotation}

Tissue areas with light RNAscope staining (indicated by small brown/yellow regions in the tissue) were identified for potential use as training data. The RNAscope transcripts on a total of 144 480x480 pixel non-overlapping patches were annotated by a trained pathology scientist and an anatomical pathologist, with an overlap of 19 patches annotated by both. The experts annotated 113 and 50 patches respectively.

\begin{figure}[htbp]
    \centering
    \includegraphics[width=0.65\textwidth]{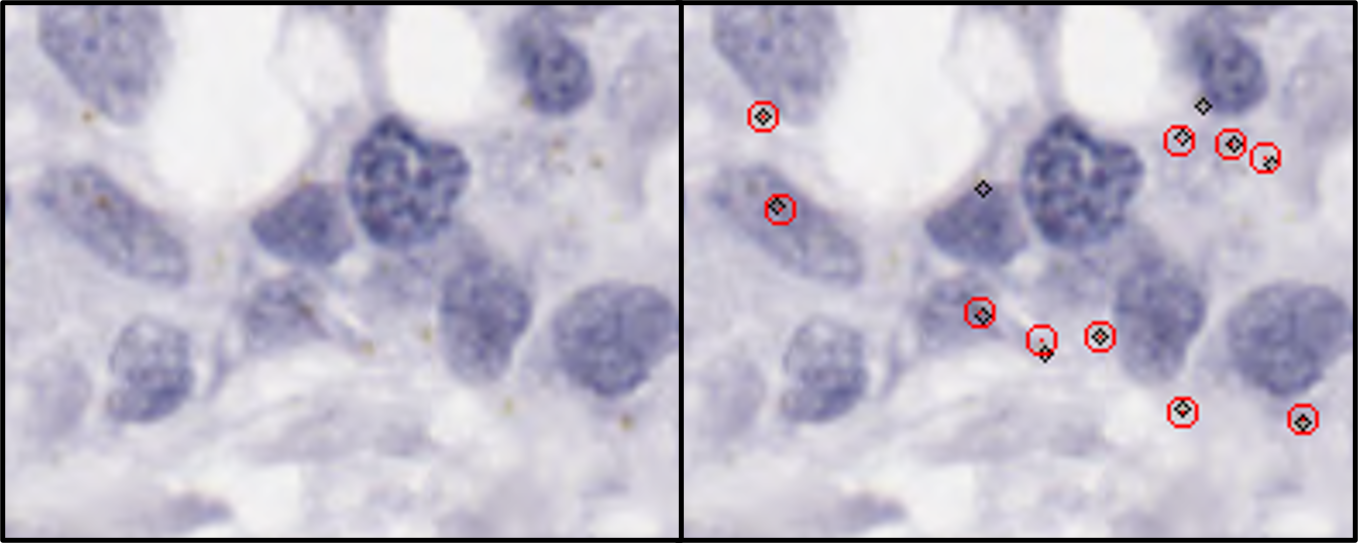}
    \caption{A visualization of the area for a potential true positive match (within 5 pixels). A clean image of the area is shown on the left, with an annotated version of the same image on the right. The first set of annotations are shown in red, with a circular boundary showing the maximum distance that could be considered a match. The second set of annotations are shown as black diamonds.}
    \label{fig:5px_radius}
\end{figure}

An agreement test was conducted on the 19 patches of tissue that were annotated with RNAscope transcript positions by both experts. This was done to provide a baseline evaluation metric. Since each set of expert annotations were represented by a set of coordinates, a method was needed to provide a reasonable agreement metric. Simply assessing exact matches of coordinates would lead to very low agreement. Therefore, pairs of annotations from each expert that were less than 5 pixels from each other were counted as true positives (as shown in figure \ref{fig:5px_radius}). Each annotation could only be paired to one annotation from the other expert. 5 pixels was chosen as the threshold value because the visibly stained portion of an RNAscope transcript generally has a diameter of up to 5 pixels. The $F_1$-score could then be calculated to evaluate the inter-rater agreement.

The $F_1$-score, representing inter-rater agreement, was found to be 0.596, with the recall of one annotator being 0.530 and the other being 0.682. It is important to note that a relatively low score is expected for this problem because each RNAscope transcript needs to not only be correctly recognized, but also detected at nearly the same position. The type of tissue staining used also causes additional difficulties, as discussed in section \ref{section:difficulties}.

\subsection{Candidate Selection}

To reduce computational load, a pre-processing step is taken to reduce the number of candidate pixels. It makes use of the colour information in the image, but cannot be made very sensitive due to the inconsistent nature of chromogenic RNAscope staining alongside haematoxylin.

Firstly, colour deconvolution (as described in \cite{deconv}) is used to separate the RNAscope (brown) stain colour into its own image. Then, the RNAscope image is Gaussian blurred using a 5x5 pixel circular kernel. Next, it is thresholded using a histogram-based method to keep only dark areas that are likely to represent RNAscope staining. A histogram for the intensities up to 250 (out of 255) is constructed. Values above 250 are excluded to remove data from the white background (non-tissue) areas. Then, the peak value is found and used as the threshold value, as this corresponds to the base tissue intensity. The remaining regions after thresholding are of lower value (indicating darker staining) than the threshold value.  If more than 50\% of the tissue still remains, the threshold value is decreased to make it more selective. A threshold decrease of 8 was found to work well. The detected RNAscope regions are separated using the Suzuki contour detection algorithm \cite{suzuki} to find each region contour. Using the grayscale representation of the original image with the same 5x5 pixel Gaussian blur applied, very dark (heavily stained) areas are found by applying a binary threshold to keep only areas with lower than 100 value. The Suzuki contour detection algorithm is again used to convert these regions into contour lists.

\begin{figure}[tbp]
    \centering
    \makebox[\textwidth][c]{\includegraphics[width=1.08\textwidth]{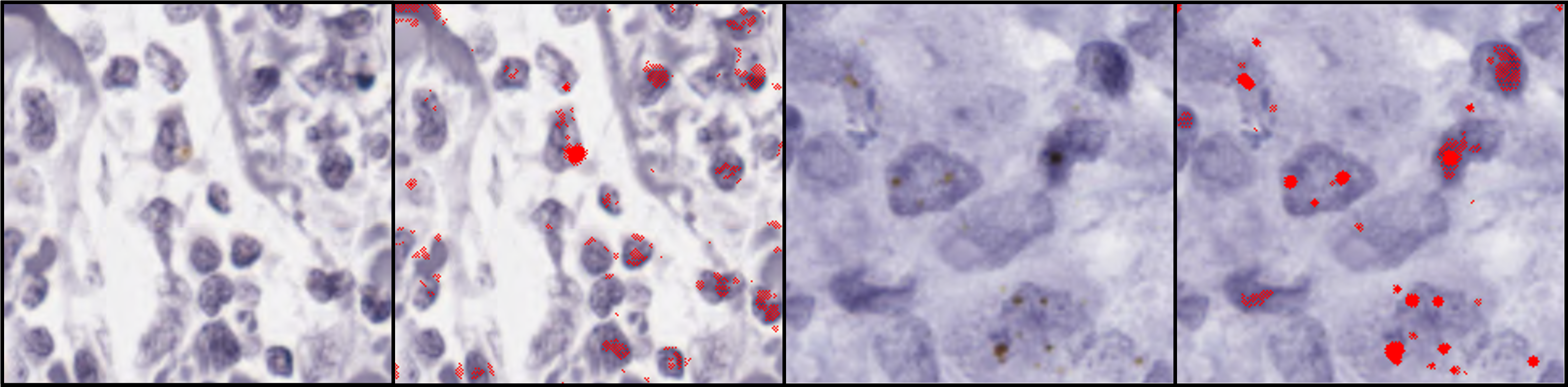}}
    \caption{An example of the candidates (red) selected from tissue, alongside their corresponding original images.}
    \label{fig:candidates}
\end{figure}

Both sets of contours are drawn onto a new, blank mask with width and height dimensions matching the original image. The background colour is the previously selected RNAscope threshold value. The RNAscope contours are filled with their original intensity values from the RNAscope image (which are darker than the threshold value). The dark region contours are filled with the RNAscope threshold value minus a small value. A value of 11 was chosen, but any value from 1 - 32 would function identically, resulting in the same candidate density of 50\%. In cases where contours overlap, the lowest value is taken. This mask is used to generate a list of candidate coordinates that may be RNAscope transcripts (dots). The list of candidates is created by repeatedly adding the coordinates of the lowest remaining value on the mask to the list of candidates. A circle centred on these coordinates (with radius according to equation \ref{eq:dot_radius}) is then drawn with its value set to the RNAscope threshold value, which prevents the area from being selected again. \textit{intensity} in the equation refers to the mask value at the candidate location, and \textit{thresh} refers to the previously selected RNAscope threshold value. This means that the radius will be 0 (producing a dot and allowing for more adjacent, clustered detections), unless the difference between the candidate location value and the RNAscope threshold value is less or equal to 32 (indicating a weaker detection). This process ends when there are no remaining values darker than the RNAscope threshold value. An example is shown in figure \ref{fig:candidates}.

\begin{equation}
 radius=floor(\frac{max(intensity - thresh + 64, 0)}{32})\label{eq:dot_radius}
\end{equation}

\subsection{Feature Extraction}

A vector of texture features is extracted for each of the candidate coordinates. The image is firstly separated into three single-channel images: grayscale, haematoxylin, and RNAscope. The haematoxylin and RNAscope channels are extracted using the previously mentioned colour deconvolution method \cite{deconv}. A large set of first order, GLRLM, GLSZM, GLCM, GLDM, and NGTDM features are extracted. For the features that can be assessed at different distances, distances of 1, 2, and 3 pixels are used. GLDM cut-offs of 0, 1, and 2 pixels are used. Each feature is assessed in both a 7x7 pixel and an 11x11 pixel window around the central pixel. Feature vectors for coordinates that are within 1 pixel of a ground-truth annotation are classified as positive, and all other coordinates as negative. The texture features are normalized. For classification, the linear support vector classifier (LSVC) was found to work well, using 1e-7 stopping tolerance, balanced class weighting, l2 penalty, C=1.0, and 1e7 max iterations. Viewing the coefficient weights for the trained LSVC showed that very few features were significant, so a reduced feature set was designed; 1548/1572 features were removed. The remaining features are first order energy and variance, NGTDM coarseness, and GLDM Large Dependence High Gray Level Emphasis (LDHGLE). NGTDM coarseness was kept over GLSZM Large Area High Gray Level Emphasis (LAHGLE, which was weighted slightly higher) as it is simpler to compute. These are only assessed at a distance of 3 pixels, and GLDM cut-off of 2 pixels. The reduced feature set extraction process runs about 10 times faster. Each classifier was trained on 115 of the annotated patches, with the remaining 29 being used for validation.

\begin{figure*}[tbp]
    \centering
    \includegraphics[width=1\textwidth]{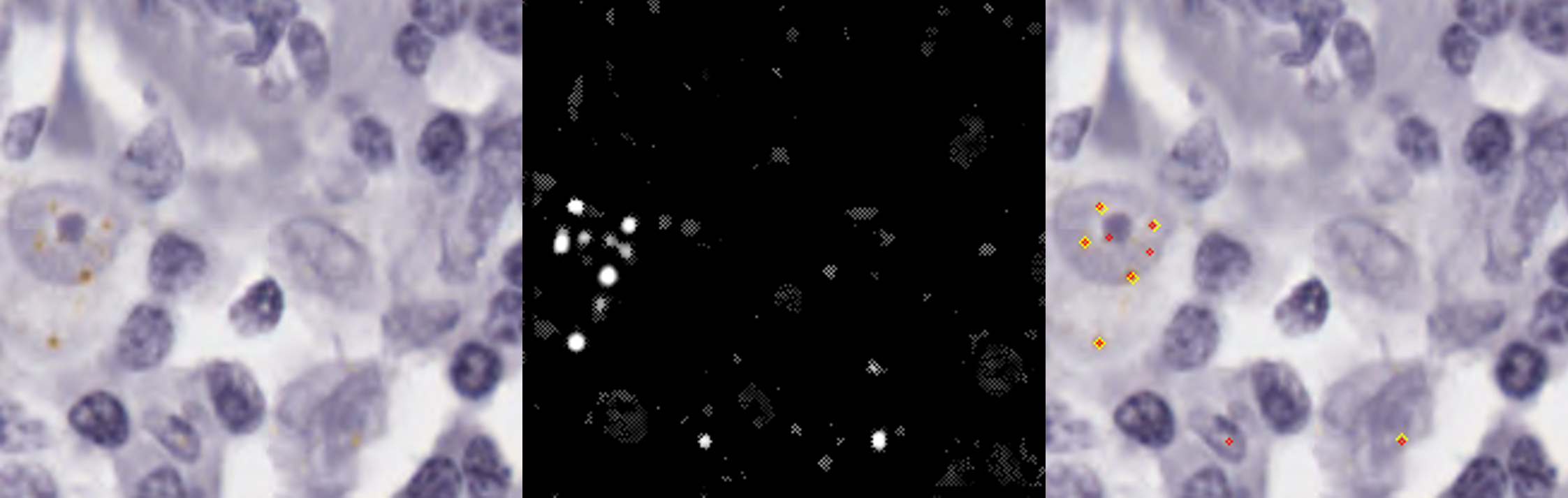}
    \caption{The output of the feature segmentation method. The leftmost image shows the input image data, the middle image shows the segmentation map, and the final image shows the detections (in red) after the watershed transform is applied. The ground truth points are shown as yellow diamonds.}
    \label{fig:feature_seg}
\end{figure*}

The prediction outputs for each feature vector are used to produce segmentation maps, which are binarized with a configurable gray threshold value. The watershed transform is used to aggregate neighboring values into a set of coordinates that can be directly compared against the ground truth coordinates. The minimum pixels in each detection (area threshold) is configurable. Figure \ref{fig:feature_seg} shows the segmentation map and final coordinates for an example patch.

\section{Results}

\label{section:feature-results}

The classifiers were evaluated against the same set of 19 patches used to assess expert inter-rater agreement. No comparison was made with other existing methods, due to their lack of output granularity and automation (as discussed in section \ref{section:lit-review}). No patches from the same tissue cores were present across both the training and validation datasets. Both the classifier using the full feature set and the classifier using the reduced feature set were evaluated. $F_1$-score (based on matching annotation pairs) was used as the evaluation metric for each classifier. The best scores for each classifier were very similar for both classifiers, with the reduced feature set classifier only performing slightly worse (0.571 $F_1$-score) than the classifier using the full feature set (0.572 $F_1$-score). Both classifiers performed similarly to the expert inter-rater agreement (0.596 $F_1$-score). Additional metrics (precision and recall) are also shown in table \ref{table:agreement_set_only}. 

\begin{table}[htbp]
\caption{Multiple accuracy measurements for both classifiers}
\centering
\begin{tabular}{|c|c | c | c|} 
 \hline
 \textbf{Classifier}& \textbf{$F_1$-score} & \textbf{Precision} & \textbf{Recall} \\ 
 \hline
 Full feature set classifier & 0.572 & 0.626 & 0.527 \\ 
 \hline
 Reduced feature set classifier & 0.571 & 0.633 & 0.521 \\
 \hline
 Experts & 0.596 & 0.682 & 0.530 \\
 \hline
\end{tabular}
\label{table:agreement_set_only}
\end{table}

\begin{figure}[htbp]
    \centering
    \includegraphics[width=0.65\textwidth]{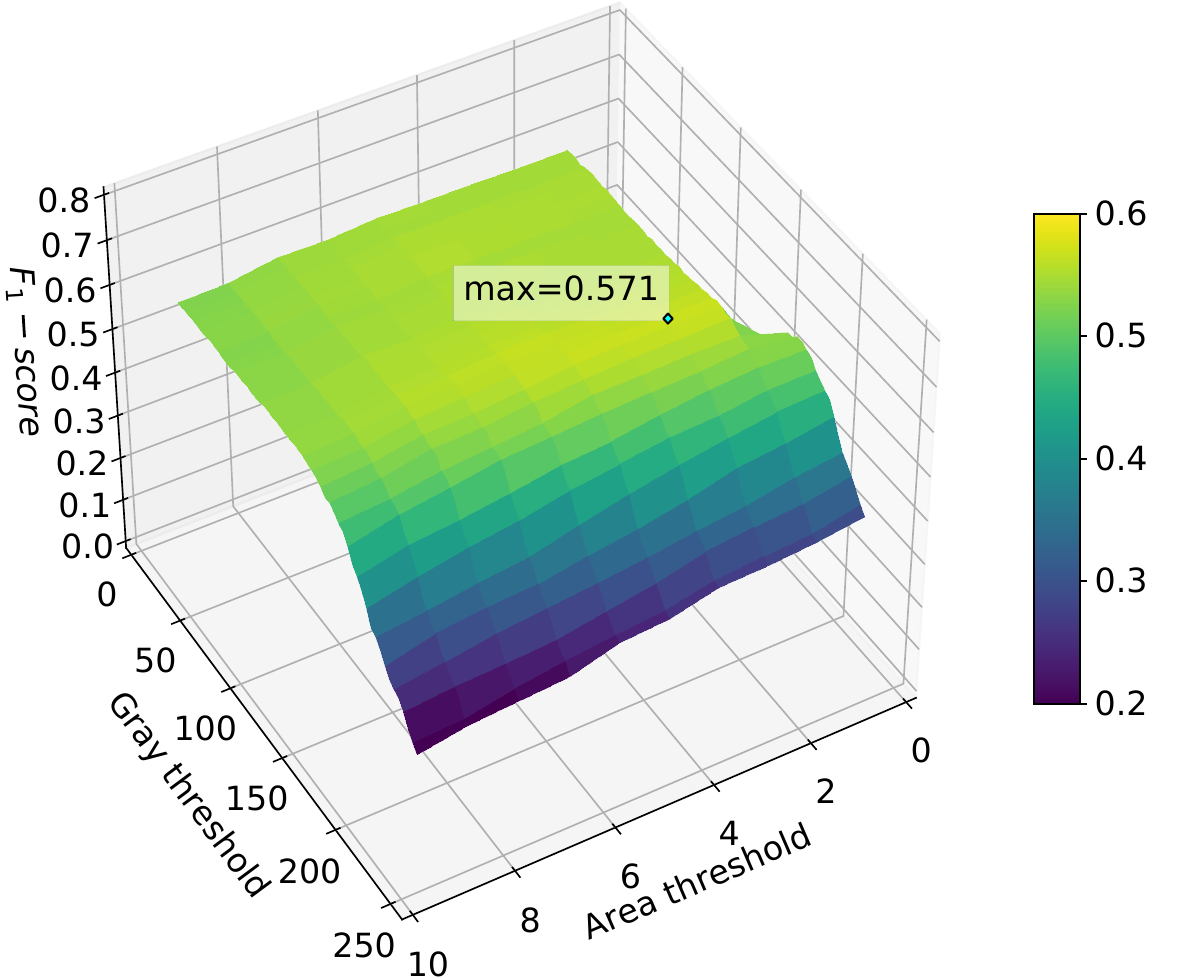}
    \caption{The $F_1$-score of the feature-based segmentation method (reduced feature set) when using differing gray thresholds and area thresholds for segmentation post-processing. The max $F_1$-score is at 2 pixel area threshold and gray threshold of 132.}
    \label{fig:feature_reduced_set}
\end{figure}

To illustrate the segmentation characteristics and hyperparameter (gray threshold and area threshold) selection process, a surface plot showing the $F_1$-score at each gray threshold and area threshold is shown in figure \ref{fig:feature_reduced_set}. Since both classifiers produced very similar surfaces, only the plot for the reduced feature set classifier is shown. The plot displays good $F_1$-scores at most thresholds, only dropping significantly at gray thresholds higher than 150. This means that both classifiers are resilient to small changes in the RNAscope detection size on the segmentation map, and to the intensity of these detections. This also implies that the feature extraction is stable, producing coherent results for neighboring pixels.

\subsection{Feature Analysis}

\begin{figure}[htbp]
    \makebox[\textwidth][c]{\includegraphics[width=1.1\textwidth]{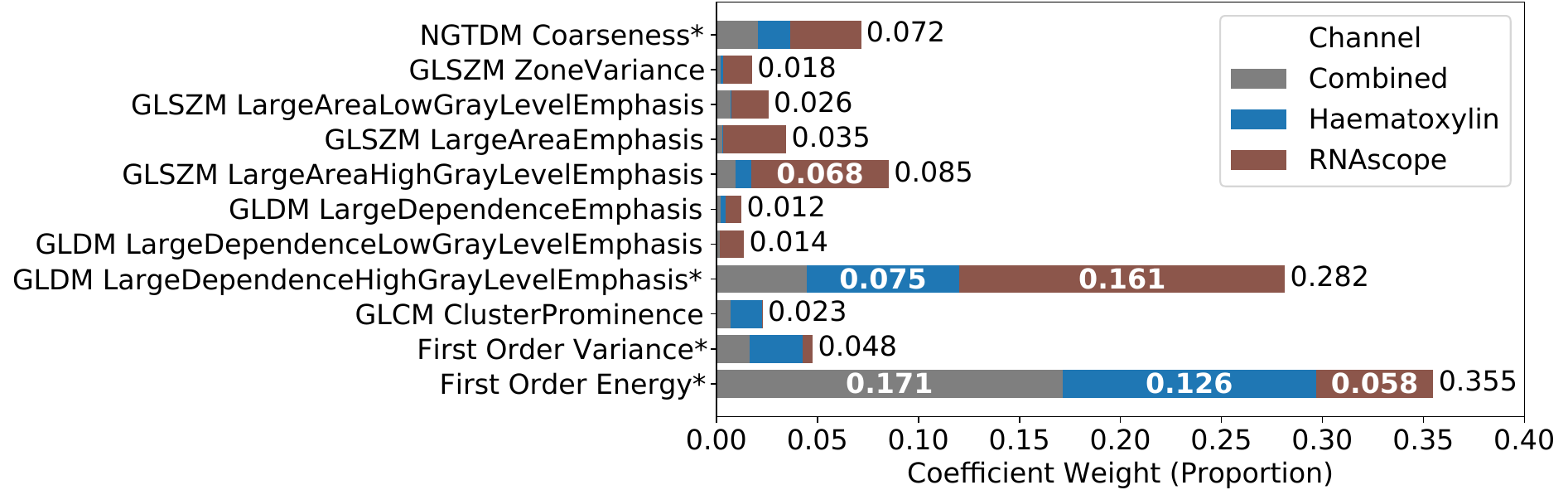}}
    \caption{A breakdown of the coefficient weight contribution of each colour channel to each feature type in the full feature set LSVC. Categories with weight smaller than 0.01 were excluded. Subcategories with weight smaller than 0.05 were not labelled. Categories which were kept for the reduced feature classifier are labelled with an asterisk.}
    \label{fig:channel_by_feature}
\end{figure}

Of all of the features extracted, few are needed to produce an accurate segmentation map. This is evidenced by the fact that reducing the feature set from 1572 to 24 features only decreased the classification $F_1$-score from 0.572 to 0.571. Figure \ref{fig:channel_by_feature} shows the relative coefficient weights of each category of features, and the weighting of each colour channel. Weighting for a feature is aggregated across all parameters (such as distance, cutoff) measured for that feature.

The feature with highest weighting is the first order energy, which is simply the magnitude of pixel values within the area. It accounts for 35.5\% of the total coefficient weight magnitude. For the smaller 7x7 pixel window, lower gray and RNAscope channel energy values (indicating darker staining) were positively correlated with RNAscope detection; whereas, lower haematoxylin values (indicating denser, darker tissue) were negatively correlated. For the larger 11x11 pixel window, the opposite correlation (albeit weaker) occurred for each of the three channels. This makes sense as the 11x11 pixel window would mostly consist of the tissue surrounding the transcript, not just the transcript itself.

The second highest weighted feature is the GLDM LDHGLE, which accounts for 28.2\% of the total weight. The LDHGLE measures the occurrence of high (light) gray valued regions with homogeneous texture. This is positively correlated with RNAscope detection on the RNAscope channel at window size 11x11 pixels. This positive correlation at the larger window size is likely because the immediate area surrounding an RNAscope transcript generally contains little to no RNAscope staining. The third highest weighted feature, GLSZM LAHGLE, indicates large connected areas of high gray value, and carries the same positive correlation on the RNAscope channel at window size 11x11 pixels. Given that these two features are so similar, including both would be largely redundant.

The fourth highest weighted feature is the NGTDM coarseness, which accounts for 7.2\% of the total weight. It assesses the average difference of gray values from their neighbors within a small (1-3 pixel) radius, or the spatial rate of change. Lower rates of change on the RNAscope and gray channels correlate with transcript detection, whereas the opposite association exists on the haematoxylin channel. This could be because a dot with smooth edges would produce low coarseness values on the RNAscope channel, and the haematoxylin channel will naturally be coarse in the areas near nuclei, where RNA is primarily found.

No GLRLM or GLCM features were highly weighted. Since GLRLM assesses horizontal runs of same-valued pixels, the noisy and two dimensional nature of slide image data would reduce its suitability. Given that GLCM assesses co-occurrence of gray values at small, fixed offsets, it does not seem well suited for detecting simple dot structures atop heterogeneous tissue.

\section{Conclusion}
In this paper, the viability of gray level texture features for segmentation of chromogenic dye RNAscope was investigated. A linear support vector classifier using a variety of these features was developed, and subsequently pruned to only include the most significant features. It performed similarly to two experts who annotated data for this study, achieving an $F_1$-score of 0.571 for identifying closely matching RNAscope transcript coordinates. The baseline expert $F_1$-score for the same task was 0.596. Feature analysis revealed a small set of gray level features well suited to segmentation of chromogenic dots (RNAscope) from histological tissue specimen slide images. Automated methods for chromogenic RNAscope segmentation and subsequent quantification are crucial for the usability of RNAscope in routine pathology workflows, and could greatly aid in cancer diagnosis with further development. For future work, more nuanced deep learning methods will be investigated because they could potentially better handle the tissue heterogeneity that makes this segmentation task difficult.

\section*{Acknowledgment}
This study used data supplied by Associate Professor Logan Walker, Dr. Arthur Morley-Bunker, and Dr. George Wiggins from the University of Otago, Christchurch.

The RNAscope transcripts on a total of 144 image patches were annotated by Dr. Arthur Morley-Bunker (a trained pathology scientist) and Dr. Gavin Harris (an anatomical pathologist) for use in this study.

We wish to thank Heather Thorne, Eveline Niedermayr, Sharon Guo, all the kConFab research nurses and staff, the heads and staff of the Family Cancer Clinics, and the Clinical Follow Up Study (which has received funding from the NHMRC, the National Breast Cancer Foundation, Cancer Australia, and the National Institute of Health (USA)) for their contributions to this resource, and the many families who contribute to kConFab. kConFab is supported by a grant from the National Breast Cancer Foundation, and previously by the National Health and Medical Research Council (NHMRC), the Queensland Cancer Fund, the Cancer Councils of New South Wales, Victoria, Tasmania and South Australia, and the Cancer Foundation of Western Australia.

\bibliographystyle{spmpsci}
\bibliography{references}

\begin{thebibliography}{10}
\providecommand{\url}[1]{{#1}}
\providecommand{\urlprefix}{URL }
\expandafter\ifx\csname urlstyle\endcsname\relax
  \providecommand{\doi}[1]{DOI~\discretionary{}{}{}#1}\else
  \providecommand{\doi}{DOI~\discretionary{}{}{}\begingroup
  \urlstyle{rm}\Url}\fi

\bibitem{NGTDM}
Amadasun, M., King, R.: Textural features corresponding to textural properties.
\newblock IEEE Transactions on Systems, Man, and Cybernetics \textbf{19}(5),
  1264--1274 (1989).
\newblock \doi{10.1109/21.44046}

\bibitem{bc-increasing}
Arnold, M., Morgan, E., Rumgay, H., Mafra, A., Singh, D., Laversanne, M.,
  Vignat, J., Gralow, J.R., Cardoso, F., Siesling, S., Soerjomataram, I.:
  Current and future burden of breast cancer: Global statistics for 2020 and
  2040.
\newblock The Breast \textbf{66}, 15--23 (2022).
\newblock \doi{10.1016/j.breast.2022.08.010}

\bibitem{gray-level-used}
Chowdhary, C.L., Acharjya, D.: Segmentation and feature extraction in medical
  imaging: A systematic review.
\newblock Procedia Computer Science \textbf{167}, 26--36 (2020).
\newblock \doi{https://doi.org/10.1016/j.procs.2020.03.179}.
\newblock International Conference on Computational Intelligence and Data
  Science

\bibitem{GLRLM}
Galloway, M.M.: Texture analysis using gray level run lengths.
\newblock Computer Graphics and Image Processing \textbf{4}(2), 172--179
  (1975).
\newblock \doi{https://doi.org/10.1016/S0146-664X(75)80008-6}

\bibitem{whole-slide-imaging}
Ghaznavi, F., Evans, A., Madabhushi, A., Feldman, M.: Digital imaging in
  pathology: Whole-slide imaging and beyond.
\newblock Annual Review of Pathology: Mechanisms of Disease \textbf{8}(1),
  331--359 (2013).
\newblock \doi{10.1146/annurev-pathol-011811-120902}

\bibitem{availability-tech}
Graschew, G., Roelofs, T.A., Rakowsky, S., Schlag, P.M.: E-health and
  telemedicine.
\newblock International Journal of Computer Assisted Radiology and Surgery
  \textbf{1}(1), 119--135 (2006).
\newblock \doi{10.1007/s11548-006-0012-1}

\bibitem{pyradiomics}
van Griethuysen, J.J., Fedorov, A., Parmar, C., Hosny, A., Aucoin, N., Narayan,
  V., Beets-Tan, R.G., Fillion-Robin, J.C., Pieper, S., Aerts, H.J.:
  {Computational Radiomics System to Decode the Radiographic Phenotype}.
\newblock Cancer Research \textbf{77}(21), e104--e107 (2017).
\newblock \doi{10.1158/0008-5472.CAN-17-0339}

\bibitem{GLCM}
Haralick, R.M., Shanmugam, K., Dinstein, I.: Textural features for image
  classification.
\newblock {IEEE} Transactions on Systems, Man, and Cybernetics
  \textbf{{SMC}-3}(6), 610--621 (1973).
\newblock \doi{10.1109/tsmc.1973.4309314}

\bibitem{MORLEYBUNKER2021151765}
Morley-Bunker, A.E., Wiggins, G.A., Currie, M.J., Morrin, H.R., Whitehead,
  M.R., Eglinton, T., Pearson, J., Walker, L.C.: Rnascope compatibility with
  image analysis platforms for the quantification of tissue-based colorectal
  cancer biomarkers in archival formalin-fixed paraffin-embedded tissue.
\newblock Acta Histochemica \textbf{123}(6), 151,765 (2021).
\newblock \doi{https://doi.org/10.1016/j.acthis.2021.151765}

\bibitem{pathologist_shortage}
Rozario, S.Y., Sarkar, M., Farlie, M.K., Lazarus, M.D.: Responding to the
  healthcare workforce shortage: A scoping review exploring anatomical
  pathologists' professional identities over time.
\newblock Anatomical Sciences Education \textbf{n/a}(n/a) (2023).
\newblock \doi{https://doi.org/10.1002/ase.2260}

\bibitem{deconv}
Ruifrok, A.C., Johnston, D.A.: {{Q}uantification of histochemical staining by
  color deconvolution}.
\newblock Anal Quant Cytol Histol \textbf{23}(4), 291--299 (2001)

\bibitem{GLDM}
Sun, C., Wee, W.G.: Neighboring gray level dependence matrix for texture
  classification.
\newblock Computer Vision, Graphics, and Image Processing \textbf{23}(3),
  341--352 (1983).
\newblock \doi{https://doi.org/10.1016/0734-189X(83)90032-4}

\bibitem{suzuki}
Suzuki, S., be, K.: Topological structural analysis of digitized binary images
  by border following.
\newblock Computer Vision, Graphics, and Image Processing \textbf{30}(1), 32 --
  46 (1985).
\newblock \doi{https://doi.org/10.1016/0734-189X(85)90016-7}

\bibitem{GLSZM}
Thibault, G., FERTIL, B., Navarro, C., Pereira, S., Lévy, N., Sequeira, J.,
  Mari, J.L.: Texture indexes and gray level size zone matrix application to
  cell nuclei classification.
\newblock In: International Conference on Pattern Recognition and Information
  Processing (PRIP’09), pp. 140--145 (2009)

\bibitem{rnascope}
Wang, F., Flanagan, J., Su, N., Wang, L.C., Bui, S., Nielson, A., Wu, X., Vo,
  H.T., Ma, X.J., Luo, Y.: Rnascope: a novel in situ rna analysis platform for
  formalin-fixed, paraffin-embedded tissues.
\newblock The Journal of molecular diagnostics : JMD \textbf{14}(1), 22--29
  (2012).
\newblock \doi{10.1016/j.jmoldx.2011.08.002}

\end{thebibliography}

\end{document}